\documentclass[letterpaper, 10 pt, conference]{ieeeconf}  

\IEEEoverridecommandlockouts                              

\usepackage{graphicx}
\usepackage[mathscr]{eucal}
\usepackage{CJK}
\usepackage[top=2cm, bottom=2cm, left=2cm, right=2cm]{geometry}
\usepackage{algorithm}
\usepackage{algorithmicx}
\usepackage{algpseudocode}
\usepackage{amsmath}
\usepackage{bm}
\usepackage{amsfonts,amssymb}
\usepackage{booktabs}
\usepackage{multirow}
\usepackage{xcolor} 
\usepackage{cite}

\title{\LARGE \bf
A Two-Stage Lightweight Framework for Efficient Land-Air Bimodal Robot Autonomous Navigation}

\author{Yongjie Li$^{1,2,\dagger}$, Zhou Liu$^{2,\dagger}$, Wenshuai Yu$^{1}$ \textit{Member,IEEE}, Zhangji Lu$^{1}$, Chenyang Wang$^{2}$, \\Fei Yu$^{2}$ \textit{Fellow,IEEE} and Qingquan Li$^{1,2,*}$
\thanks{* denotes corresponding author.}
\thanks{$\dagger$ These authors contributed equally to this work.}%
\thanks{$^{1}$Yongjie Li, Wenshuai Yu, Zhangji Lu and Qingquan Li are with Shenzhen University}%
\thanks{$^{2}$Yongjie Li, Zhou Liu, Chenyang Wang, Fei Yu and Qingquan Li are with the Guangdong Laboratory of Artificial Intelligence and Digital Economy (Shenzhen), Shenzhen 518107, China}%
}

\begin{document}

\maketitle
\thispagestyle{empty}
\pagestyle{empty}

\begin{abstract}

Land-air bimodal robots (LABR) are gaining attention for autonomous navigation, combining high mobility from aerial vehicles with long endurance from ground vehicles. However, existing LABR navigation methods are limited by suboptimal trajectories from mapping-based approaches and the excessive computational demands of learning-based methods. To address this, we propose a two-stage lightweight framework that integrates global key points prediction with local trajectory refinement to generate efficient and reachable trajectories. In the first stage, the Global Key points Prediction Network (GKPN) was used to generate a hybrid land-air keypoint path. The GKPN includes a Sobel Perception Network (SPN) for improved obstacle detection and a Lightweight Attention Planning Network (LAPN) to improves predictive ability by capturing contextual information. In the second stage, the global path is segmented based on predicted key points and refined using a mapping-based planner to create smooth, collision-free trajectories. Experiments conducted on our LABR platform show that our framework reduces network parameters by 14\% and energy consumption during land-air transitions by 35\% compared to existing approaches. The framework achieves real-time navigation without GPU acceleration and enables zero-shot transfer from simulation to reality during deployment.

\end{abstract}

\section{INTRODUCTION}

As the Low-altitude economy continues to develop, land-air bimodal robots (LABR) have gained significant attention for their maneuverability and durability. These robots autonomously switch between land and air motion modes, effectively overcoming the limitations of ground-only navigation, which prevents reaching goals in mid-air. Moreover, they avoid the energy consumption associated with prolonged air travel. This significantly improves LABR reachability and endurance in complex environments, making them particularly effective in applications such as search and rescue missions\cite{rescue}, reconnaissance missions\cite{Liu}, and item deliveries\cite{Liu2}.

\begin{figure}[thpb]

      \centering
    \includegraphics[width=1\linewidth]{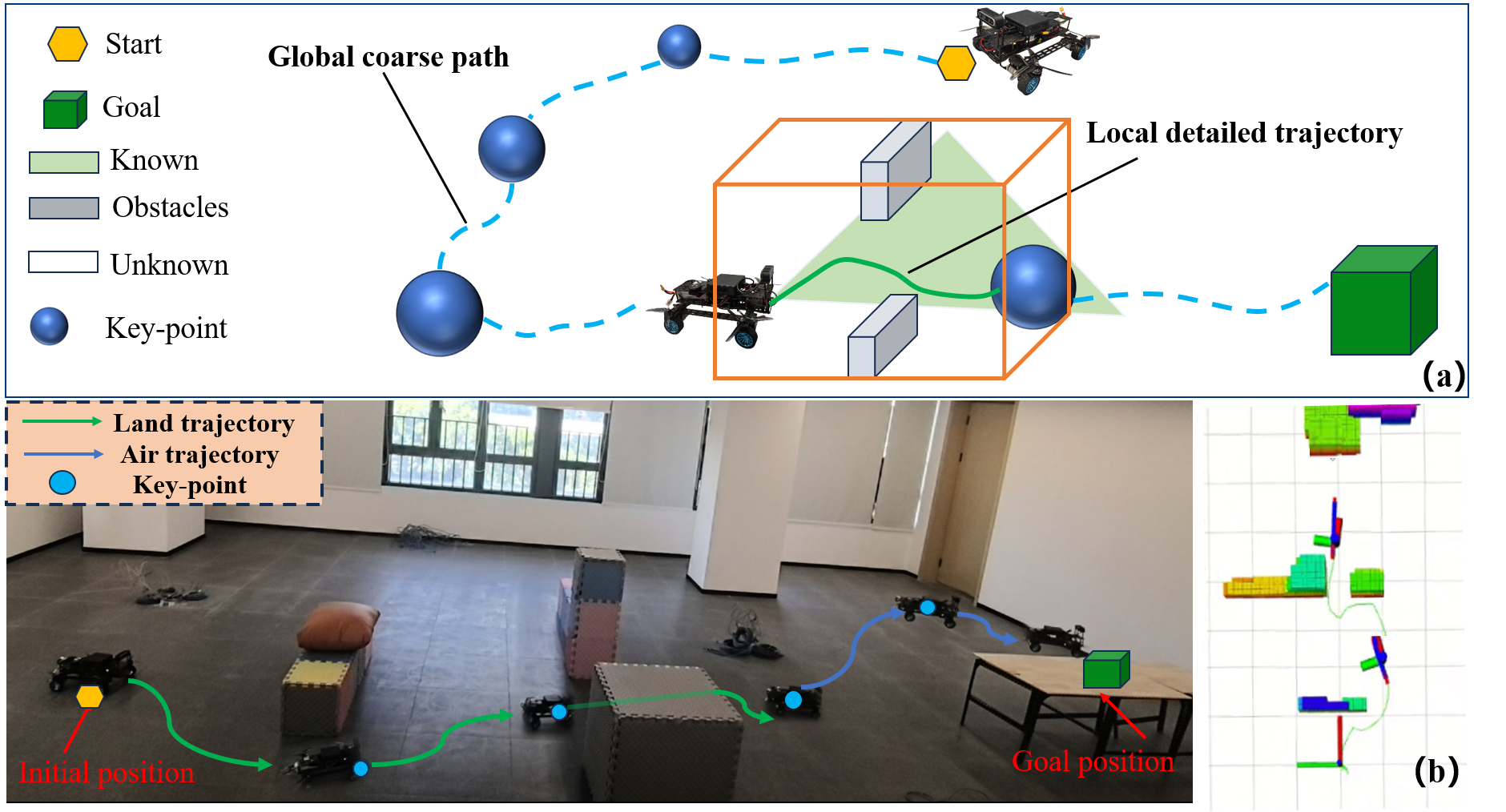}
      \caption{(a) Illustration of our navigation framework. Inside the local planning horizon (orange box), a local detailed trajectory is computed. At the global scale, the global key points are predicted and the coarse path is obtained. (b) Our method is used in real-world scenarios where land-air switching is required.}
      \label{figure1}
      \vspace{-0.6cm}
\end{figure}

Existing mapping-based research\cite{tabv,hybrid,Ego} has confirmed the navigation capabilities of LABR in simple environments. These methods rely on onboard sensors to construct and update environmental maps, thereby facilitating effective path planning. However, traditional mapping-based methods, while capable of using sensor data to update visible obstacle information, may still pose potential collision risks due to the limited visibility of local maps\cite{mapdr}. Furthermore, the lack of terrain feature data in unknown areas can result in overly conservative trajectory choices during long-distance planning, leading to unnecessary energy consumption. In contrast, recent studies\cite{RL,AGRNav,RLdrawback} have increasingly adopted learning-based methods for path planning. AGRNav\cite{AGRNav} proposes a semantic scene completion network for predicting obstacles. Although this approach reduces collision probabilities, AGRNav introduces additional computational overhead, increasing onboard processing requirements.

Notably, iPlanner\cite{iPlanner} introduces an efficient approach based on imperative learning, which inputs individual depth measurements into a lightweight network to generate key points path towards the goal. Then, iPlanner uses cubic-spline fitting to create a smooth trajectory between key points. The generation of key points path improves the efficiency of global planning execution. However, it lacks a three-dimensional motion coordination mechanism, making it difficult to meet the demands of land-air bimodal navigation. Moreover, iPlanner only relies on depth information and lacks more detailed edge features, resulting in a lack of robustness in obstacle-dense scenarios. Consequently, a pivotal challenge persists: \textbf{\textit{how to utilize limited perceptual and computational resources to balance land and air navigation, optimizing energy consumption and ensuring reachability to goal regions?}}


To address the above challenges, we propose a two-stage lightweight framework that integrates the global understanding provided by learning-based methods\cite{iPlanner} with the precision of mapping-based local planning\cite{Ego}, as shown in Fig. \ref{figure1}. In the first stage, a Global Key points Prediction Network (GKPN) was trained to generate a hierarchical key points path (i.e., land and air points), providing a coarse global path toward the goal. The GKPN architecture integrates two components: the Sobel Perception Network (SPN), which leverages edge information to improve obstacle detection, and the Lightweight Attention Planning Network (LAPN), which captures rich contextual information and features of occluded areas. In the second stage, the global path is divided into several local segments based on the generated key points, where traditional mapping-based planning refines the detailed local trajectory. Through this two-stage integration, the approach enables real-time coordination between global guidance and local obstacle avoidance, delivering an efficient and reliable solution for autonomous navigation.

The main contributions of this paper are as follows:
\begin{itemize}
        \item A two-stage lightweight framework integrating global prediction with local optimization was proposed, achieving 35\% energy consumption reduction compared to mapping-based method in land-air switching scenarios while ensuring reachability to goal.
	\item A Global Key-points Prediction Network was developed to generate land-air bimodal navigation key points, featuring a lightweight architecture that reduced network parameters by 14\% compared to the baseline network\cite{iPlanner}.
        \item To the best of our knowledge, this constitutes the first implementation of learning-based method on GPU-free LABR platforms, successfully accomplishing zero-shot transfer from simulation to real-world deployment.
\end{itemize}

\section{RELATED WORK}

\subsection{Mapping-Based Navigation}
Mapping-based methods\cite{robust,fiesta,TNS,efp} form the foundation of autonomous navigation, utilizing onboard sensors like LiDARs and cameras to construct and update environmental maps for precise path planning. For example, \cite{TNS} extracts terrain features from RGB images and 3D point clouds to adapt to the environment and update terrain information. \cite{efp} employed the UFOMap algorithm to represent the entire environment and reduce map construction time, leveraging accurate representations and hierarchical boundary structures for rapid boundary extraction. However, these methods are limited by sensor detection range, lack robustness against noise in real-world tasks, and require frequent map updates, leading to increased overall latency\cite{mapdr}.

\subsection{Learning-Based Navigation}

Recently, learning-based methods have emerged as a promising solution for autonomous navigation. These methods are trained using reinforcement learning (RL)\cite{RL2,RL3,Hoeller,NAMO3}, imitation learning (IL)\cite{mapdr}, or very recently via imperative learning (ImpL)\cite{iPlanner}. For example, \cite{Hoeller} trained an obstacle avoidance strategy using reinforcement learning by fusing image sequences with the current camera trajectory. \cite{mapdr} proposed using imitation learning to train a neural network that directly maps noisy depth images to collision-free trajectories. Notably, \cite{iPlanner} introduced  a novel imperative learning path planning method that combines bi-level optimization to improve planning efficiency while reducing the reliance on large amounts of labeled data. While these methods accelerate convergence compared to mapping-based approaches, they still require large amounts of real-world data for training to bridge the gap with reality.

\subsection{Land-air Bimodal Robot Navigation}

To address the challenge of autonomous navigation in land-air switching scenarios, land-air bimodal robots (LABR) have gained significant attention. Some studies focus on designing lighter mechanical structures\cite{multimodal,multimodal2,skywalker}. For example, Skywalker\cite{skywalker} uses four independent propellers and a centrally driven wheel to achieve bimodal movement. Meanwhile, researchers have made significant breakthroughs in developing energy-efficient navigation frameworks\cite{AGRNav,tabv,hybrid}. For instance, \cite{tabv} adopted a hierarchical motion planning approach, coupled with a unified motion controller capable of dynamically adjusting energy consumption. \cite{hybrid} used differential flatness mapping to implement a unified planner that enables the robot to autonomously switch between land-rolling and air-flying modes. Although these methods have been successful, their real-time performance in real-world environments remains constrained by onboard computational power.

\section{PROBLEM DEFINITION}

Our objective is to achieve autonomous navigation towards the goal in both land and air scenes. In addition, the energy efficiency and feasibility of the trajectory are ensured during the navigation process. Mathematically, define $\mathcal{S} \subset \mathbb{R}^{3}$ as the workspace for the robot to navigate, let the subset $\mathcal{S}_{obs} \subset \mathcal{S}$ represent the obstacles in the space that the robot cannot traverse through. At each time step $t$, the robot is initialized at a random position $\mathbf{p} _{t}^{R} \in \mathbb{R}^{3}$ in the environment and receives the $W\times H$ depth image $\mathcal{D}_{t}\in \mathbb{R} ^{W\times H} $, the goal position $\mathbf{p} _{t}^{G} \in \mathbb{R}^{3}$ as inputs. The Global Key points Prediction Network (GKPN) then predicts a key points path towards the goal, denoted as $\mathcal{K}_t=GKPN(\mathcal{D}_{t},\mathbf{p} _{t}^{G},\theta)$, which consists of $n$ key points. Here, $\theta$ is the weight of the neural network. Subsequently, during training, we
use cubic-spline \cite{cubic} to interpolate $m$ intermediate points in between every two key points in $\mathcal{K}_t$ and generates a dynamically feasible trajectory ${\tau}_t \in \mathbb{R}^{(mn+1)\times 3}$, while avoiding obstacles 
 $\mathcal{S}_{obs}$. Additionally, we define a cost function $\mathcal{C}$ and fear loss $\mathcal{F}$, and the overall task loss $\mathcal{L}=\mathcal{C}+\mathcal{F}$ is backpropagated through the network to update the parameters $\theta$.

\begin{figure}[thpb]
      \centering
    \includegraphics[width=1\linewidth]{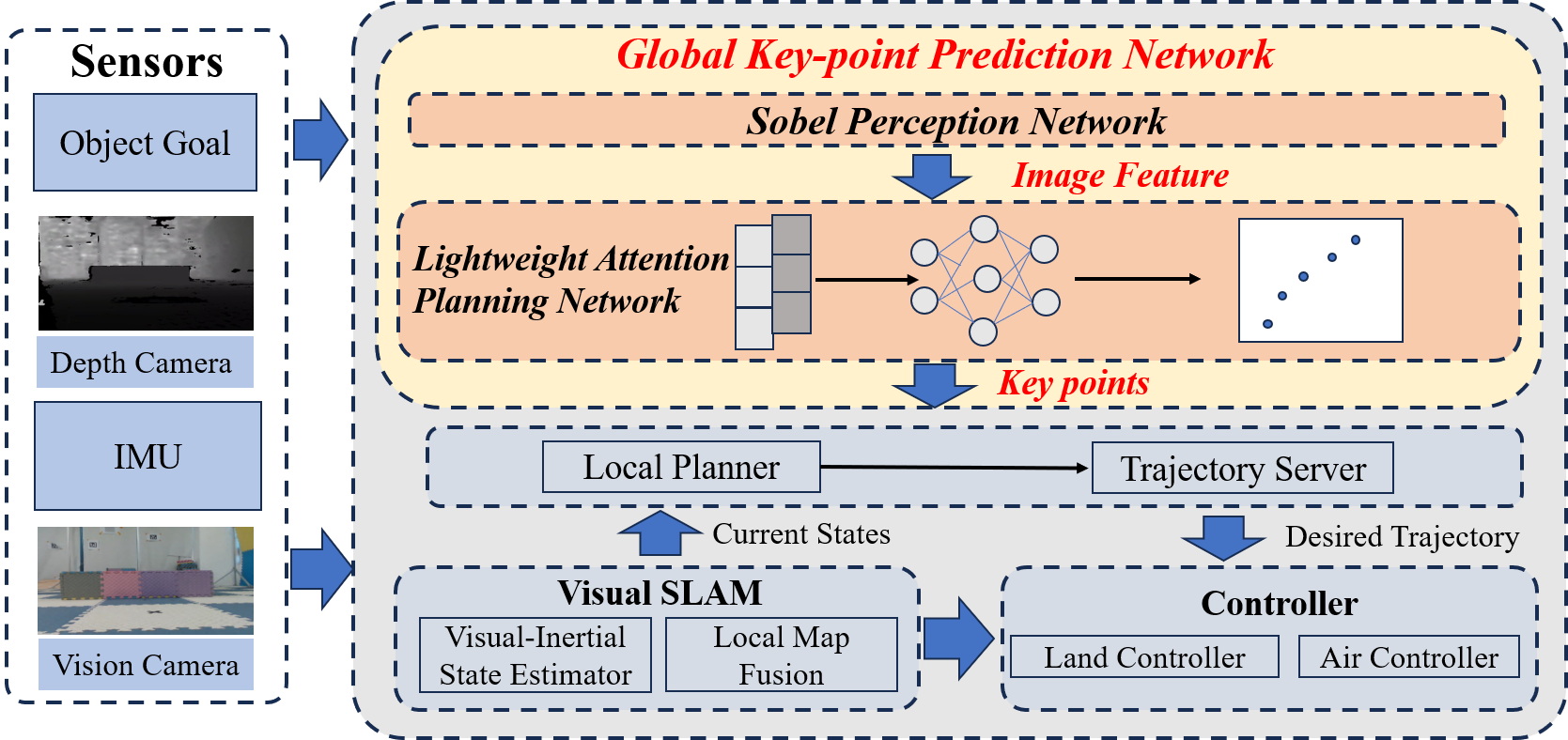}
      \caption{The flowchart of lightweight navigation framework.}
      \label{Overview}
\end{figure}

\begin{figure*}[thpb]
      \centering
    \includegraphics[width=0.85\linewidth]{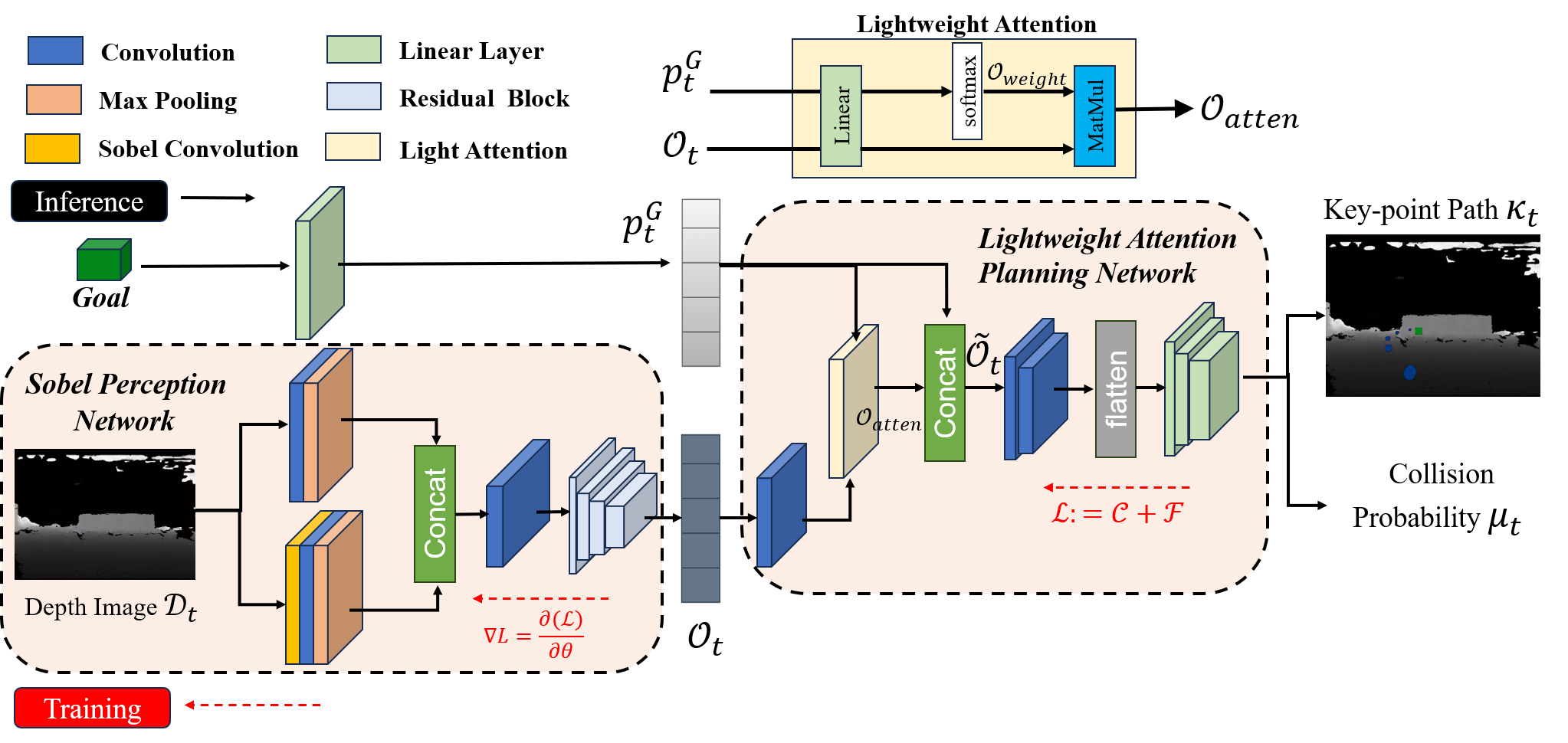}
      \caption{The Global Key points Prediction Network Structure.}
      \label{GKPN}
\vspace{-0.4cm}
\end{figure*}

\section{METHOD}

\subsection{Overview }
The overview of the proposed lightweight navigation framework is shown in Fig. \ref{Overview}. First, visual-inertial state estimation is performed by fusing camera and IMU data through Visual Simultaneous Localization and Mapping (VSLAM). Next, the goal position and the current depth image are fed into the Global Key points Prediction Network (GKPN). In the GKPN, the Sobel Perception Network enhances the depth features by integrating edge information, while the Lightweight Attention Planning Network leverages the long-range capture ability of attention mechanisms to predict a key points path towards the goal. It is worth mentioning that since iPlanner can only generate land trajectories, in order to conform to land-air switching scenarios and to reduce memory overhead, we only generate key points during deployment instead of full trajectories. Subsequently, based on the discrete key points, the well established local planner\cite{Ego} smooths and interpolates these points to generate a collision-free trajectory. Finally, the robot was controlled to track the desired trajectory via land and air controllers. 

\subsection{Global Key points Prediction Network Structure}

We propose a Global Key points Prediction Network (GKPN) to predict the key points for reaching the goal in the scene, as shown in Fig. \ref{GKPN}. 

\subsubsection{Sobel Perception Network}

Generally, depth information is primarily used for environmental perception and obstacle detection. However, a pure depth image only provides distance information for each pixel, lacking sensitivity to object shapes, edges, and other details. This can make it difficult to identify obstacles in regions with unclear object boundaries or gradual depth changes. To address this issue, this paper proposes the Sobel Perception Network (SPN), which combines Sobel convolution with original depth data. By combining the edge-enhancing Sobel operator with depth information, our approach more effectively extracts object contours. Specifically, at each time step $t$, the robot receives a depth image $\mathcal{D}_{t}$ sampled from the space $\mathcal{S}$. In the SPN, one branch extracts image features from $\mathcal{D}_{t}$ using a traditional convolutional neural network (CNN), while the other branch extracts edge features through Sobel convolution. The outputs from both branches are then concatenated and processed by a residual block to integrate the information and obtain $\mathcal{O}_{t}$:
\begin{equation}
	\mathcal{O}_t = RB([Convpool(\mathcal{D}_t),SConvpool(\mathcal{D}_t)]),
	\label{}
\end{equation}
where $Convpool$ denotes the combination of convolutional and pooling layers, $SConvpool$ refers to the addition of Sobel convolution on top of this, $[\: ,\:]$ represents concatenation and $RB$ stands for residual block\cite{resnet}. The output $\mathcal{O}_{t}\in \mathbb{R} ^{C\times M} $, $C$ denotes the channel dimension and $M$ represents the dimension of the feature space.

\subsubsection{Lightweight Attention Planning Network}
The Lightweight Attention Planning Network receives the perception embedding $\mathcal{O}_{t}\in \mathbb{R} ^{C\times M} $ and the goal position $\mathbf{p} _{t}^{G} \in \mathbb{R}^{3}$ to identify collision-free key points towards the goal.
To improve path prediction and decision-making efficiency, a lightweight attention mechanism is introduced to emphasize goal-relevant spatial features. Compared with traditional attention mechanisms that require quadratic computational complexity due to the construction of query, key, and value matrices, the proposed method adopts a single linear mapping and feature reweighting, achieving linear complexity and significantly accelerating the planning.

Specifically, to enrich the representation of the goal position information, a linear layer is first used to map it to a high-dimensional feature embedding $\mathbf{p} _{t}^{G} \in \mathbb{R}^{C_{1}\times M}$, where $C_{1}> 3$. Meanwhile, the dimension of $\mathcal{O}_{t}$ is adjusted to match that of $\mathbf{p} _{t}^{G}$. The two are then input into the lightweight attention mechanism to obtain $\mathcal{O}_{atten}\in \mathbb{R} ^{C_{1} \times M}$:
\begin{equation}
	\mathcal{O}_{weight}=Softmax(Linear(\mathbf{p}_{t}^{G} )),
	\label{}
\end{equation}

\begin{equation}
	\mathcal{O}_{atten}=Linear(\mathcal{O}_{t}) \otimes {\mathcal{O}_{weight}}^{T}.
	\label{}
\end{equation}
Subsequently, $\mathcal{O}_{atten}$ and the expanded goal feature $\mathbf{p} _{t}^{G}$ are concatenated to form $\tilde{\mathcal{O}_{t}} \in \mathbb{R} ^{(C_{1} +C_{1} ) \times M}$. Finally, the combined features undergo a series of convolutional and linear operations to process $\tilde{\mathcal{O}_{t}}$ and predict the key points path $\mathcal{K}_t\in \mathbb{R} ^{n\times 3}$ containing $n$ key points:

\begin{equation}
	\mathcal{K}_{t}=Linear(\sigma(Conv(\tilde{\mathcal{O}_{t}}))),  
	\label{}
\end{equation}
where $Conv$ denotes the convolutional layer, $\sigma$ denotes the ReLU activation function.

\subsection{Optimization Objective and Training Loss}
\subsubsection{Trajectory Cost}
The trajectory cost includes three differentiable terms that assess the quality of the output trajectory ${\tau}_t$. The loss term is formulated as follows:
\begin{equation}
	\mathcal{C}(\tau_t)=\alpha \mathcal{C}^{\mathcal{O}} +\beta \mathcal{C}^{\mathcal{M}}+\gamma \mathcal{C}^{\mathcal{G}}+\delta \mathcal{C}^{E}.
	\label{}
\end{equation}
Here, $\alpha$, $\beta$, $\gamma$ and $\delta$ are for loss scaling. The obstacle loss $\mathcal{C}^{\mathcal{O}}$, which checks whether the trajectory collides or comes too close to obstacles, and motion loss $\mathcal{C}^{\mathcal{M}}$, which evaluates the motion smoothness of the trajectory, are the same as those used in iPlanner\cite{iPlanner}.

To ensure goal reachability, we define the goal loss $\mathcal{C}^{\mathcal{G}}$ to enforce proximity between the last key point and the goal position. A conditional computation strategy is implemented to enhance computational efficiency: when the height of the goal is lower than the robot's height, the Euclidean distance is computed only in the $x$ and $y$ dimensions; otherwise, it is computed in all three dimensions. We also introduce a log scaling to limit the influence of the goal point at large distances. The loss can be expressed as:
\begin{equation}
	gloss_{i}=\begin{cases} \left \| \mathbf{p} _{n_{(x,y)} }^{K}-\mathbf{p} _{i_{(x,y)} }^{G}\right \|_{2}  
  & \text{ if } goal_{z} < h_{R} \\
  \left \| \mathbf{p} _{n_{(x,y,z)} }^{K}-\mathbf{p} _{i_{(x,y,z)} }^{G}\right \|_{2}  & \text{ otherwise }
\end{cases},
	\label{}
\end{equation}

\begin{equation}
	\mathcal{C}^{\mathcal{G}}=\frac{1}{m\times n}\sum_{i=1}^{m\times n}\log_{}{} (gloss_{i}+1 ),
	\label{}
\end{equation}
where $\mathbf{p} _{n}^{K}$ denoting the last point of the key points path. $goal_{z}$ and $h_{R}$ represent the height of the goal and the height of the robot, respectively.

To optimize the system's endurance, the energy consumption loss $\mathcal{C}^{E}$ prioritizes the use of land mode for navigation in order to minimize flight distance and associated energy consumption. Our approach aims to make the $z$-value of the trajectory $\mathbf{p} _{i_{(z)} }^{K}$ as close as possible to the robot's height $h_{R}$, and penalizes vertical deviations above the robot's altitude. This makes the path as energy efficient as possible, suppressing unwanted altitude variations. The loss can be expressed as:

\begin{equation}
	\mathcal{C}^{E}=\frac{1}{m\times n}\sum_{i=1}^{m\times n}|\mathbf{p} _{i_{(z)} }^{K}-h_{R}|.
	\label{}
\end{equation}

\subsubsection{Fear Loss}
In addition to the trajectory cost, during training, the network predicts the collision probability $\mu_{t}$ for each trajectory to assess its risk of collision with obstacles. In some cases, the local planning strategy may get stuck in a suboptimal solution, potentially failing to navigate around obstacles effectively. In real-world deployment, only planned path with collision probability $\mu_{t}<0.5$ are output, which allows the planner to avoid falling into suboptimal solutions while still maintaining safety. The fear loss $\mathcal{F}({\tau}_t)$ is computed by binary cross entropy (BCE):
\begin{equation}
  \mathcal{F}(\tau_t)=\begin{cases} BCEloss(\mu_{t},1.0)
  &  \mathbf{p} _{i}^{K}\in \mathcal{S}_{obs} \\ BCEloss(\mu_{t},0.0)
  & \text{ otherwise }
\end{cases}.
\end{equation}

The final training loss $\mathcal{L}$ is defined as the sum of the trajectory cost $\mathcal{C}$ and the fear loss $\mathcal{F}$:
\begin{equation}
  \mathcal{L}=\mathcal{C}(\tau_t)+\mathcal{F}(\tau_t).
  \label{total}
\end{equation}


\begin{figure}[thpb]
      \centering
    \includegraphics[width=1\linewidth]{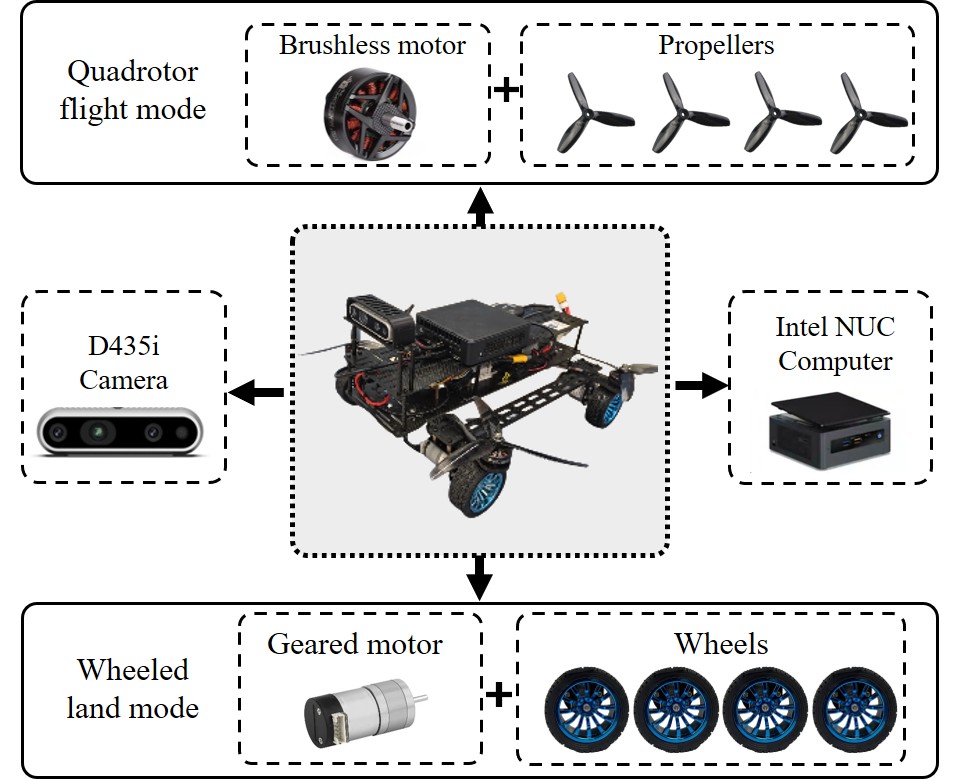}
      \caption{The structure diagram of the self-developed land-air bimodal
 robot.}
      \label{LABV}
\end{figure}

\section{EXPERIMENTS}
\subsection{Platform development of land-air bimodal robot}
The land-air bimodal robotic platform seamlessly integrated land and air navigation capabilities, as shown in Fig. \ref{LABV}. To ensure flight endurance, a carbon fiber composite frame was structurally optimized for weight reduction (total weight: 3.4 kg) with maintained integrity. The platform was equipped with essential sensors, including a D435i RGB-D camera for sensing the surrounding environment and an embedded IMU module for real-time attitude estimation. To enable autonomous navigation, an onboard computer based on NUC had been installed to meet the computing requirements of the navigation tasks designed in this project. A flight controller based on PIXHAWK and a land controller based on STM32 were implemented to execute the desired motion strategy. Power management employed a 19V 5S lithium battery, providing 15 minutes of flight endurance and 1 hour of terrestrial operation.

\subsection{Simulation Experiments}

Our method is trained using the publicly available iPlanner data set \cite{iPlanner}. The data set consists of approximately 20k depth images collected from various camera positions in the Matterport3D \cite{matterport3d} environment, as well as simulated campus and tunnel environments \cite{tunnel} using Gazebo \cite{gazebo}. To validate the effectiveness of our proposed Global Key points Prediction Network, we conducted comparative experiments with iPlanner in four different simulated environments, including indoor, garage, forest, and Matterport3D room. To ensure fairness, the experiments were conducted by only modifying the network structure while keeping the hyperparameter settings and loss calculations identical. 

\begin{table}[h]
\renewcommand{\arraystretch}{0.8}  
\caption{Ablation study of our model design choices on the four types of environments.}
\label{table1}
\centering
\begin{tabular}{ccccc}
\hline
\multirow{2}{*}{Environment} & \multicolumn{2}{c}{Method} & \multirow{2}{*}{Train. Loss} &  \multirow{2}{*}{Goal Reached} \\ 
             & SPN & LAPN  & & \\ \hline
\multirow{4}{*}{Matterport3D} & $ \times $  & $ \times $ & 3.671   & 64\% \\
                              & $ \times $ & $ \surd $  & 3.542   & 68\%\\ 
                              & $ \surd $ & $ \times $  & 3.320  & 72\% \\ 
                              & $ \surd $  & $ \surd $ & \textbf{3.221}  & \textbf{76\%}\\ \hline
\multirow{4}{*}{Forest}   & $ \times $  & $ \times $  & 0.664  & 94\%  \\
                          & $ \times $ & $ \surd $   & 0.645   &  94\%\\ 
                              & $ \surd $ & $ \times $  & 0.642  & 95\%\\ 
                           & $ \surd $  & $ \surd $ & \textbf{0.636}  & \textbf{96\%}\\ \hline
\multirow{4}{*}{Indoor}   & $ \times $  & $ \times $  & 1.588   & 86\%    \\
                          & $ \times $ & $ \surd $   &  1.457  & 84\%\\ 
                              & $ \surd $ & $ \times $  &  1.523   & 88\%\\
                           & $ \surd $  & $ \surd $ & \textbf{1.322}   & \textbf{91\%}\\ \hline
\multirow{4}{*}{Garage} & $ \times $  & $ \times $  & 0.679   & 95\%  \\
                         & $ \times $ & $ \surd $   &  0.723   & 94\%\\ 
                              & $ \surd $ & $ \times $  & 0.665   & 95\%\\
                         & $ \surd $  & $ \surd $ & \textbf{0.632}  & \textbf{96\%}\\ \hline
\end{tabular}
\end{table}

\begin{figure}[thpb]
\vspace{-0.1cm}
      \centering
    \includegraphics[width=0.9\linewidth]{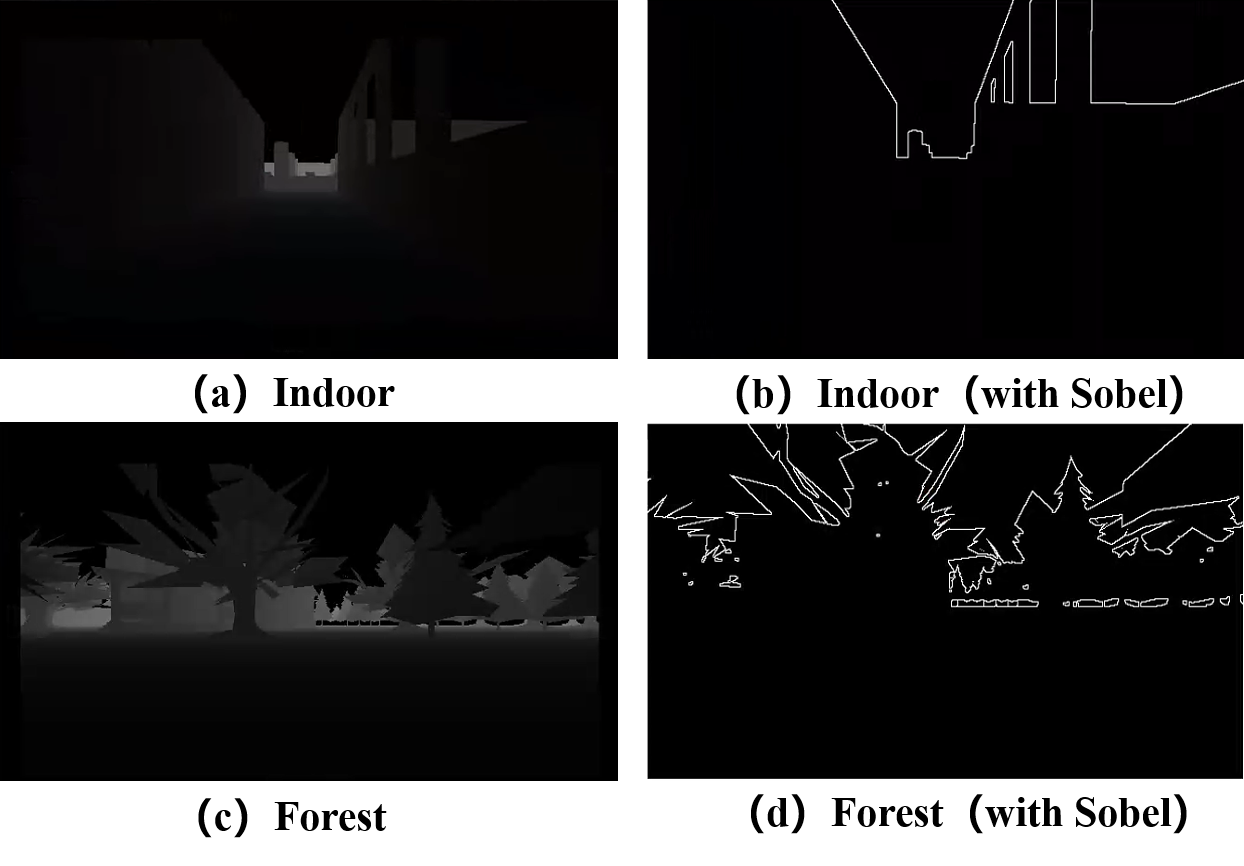}
    \vspace{-0.1cm}
      \caption{Sobel-based edge detection examples in indoor/forest environments. }
      \label{SObel}
\vspace{-0.2cm}
\end{figure}

Table \ref{table1} presents the results of the ablation study, where SPN refers to the Sobel Prediction Network, LAPN to the Lightweight Attention Planning Network, and the baseline (iPlanner without SPN and LAPN) is first row in each environment. Training loss reflects the trajectory traversability performance during training, while "Goal Reached" indicates the robot's success in reaching the goal, defined as being within 0.5m of the goal after 100 trials in each environment. In environments with simpler navigation, such as Forest and Garage, our method achieves a 2\% improvement in the goal-reached rate. In more obstacle-dense environments like Indoor and Matterport3D, compared to the baseline, our combined SPN+LAPN achieves the highest success rates (76\% in Matterport3D, and 91\% in Indoor) and lowest training losses. SPN reduces collisions by enhancing obstacle awareness, while LAPN captures long-range dependencies, effectively predicting a broader unknown area. As shown in Fig. \ref{SObel} (b) and (d), the Sobel operator effectively emphasizes object edges, compensating for the lack of shape detail in the original depth images. 

\begin{table}[h]
\renewcommand{\arraystretch}{1.2}  
\caption{Comparison of Model Parameters and FLOPs.}
\label{table2}
\centering
\begin{tabular}{ccc}
\hline
\multirow{2}{*}{Method} & \multirow{2}{*}{Params (M)} & \multirow{2}{*}{FLOPs (G)} \\ 
                        &                               &                           \\ \hline
iPlanner\cite{iPlanner}                & 54.3                   & 20.64                     \\
Only use LAPN               & 45.6                        & 16.75 \\
Only use SPN                 & 53.2                        & 21.12                      \\ 
SPN+LAPN               & 47.7                         & 17.8                      \\ \hline
\end{tabular}
\vspace{-0.2cm}
\end{table}

Table \ref{table2} compares the model parameters and computational load (FLOPs) across different methods. Our SPN+LAPN strikes an optimal balance between performance and efficiency, reducing parameters by 14\% and FLOPs by 13.8\% compared to iPlanner. The lightweight design of LAPN contributes significantly to this efficiency, while SPN’s edge extraction enhances robustness in complex scenes without excessive computational overhead. The synergy between the two ensures superior navigation capabilities across diverse environments with practical deployability.

\begin{figure}[thpb]
      \centering
    \includegraphics[width=1\linewidth]{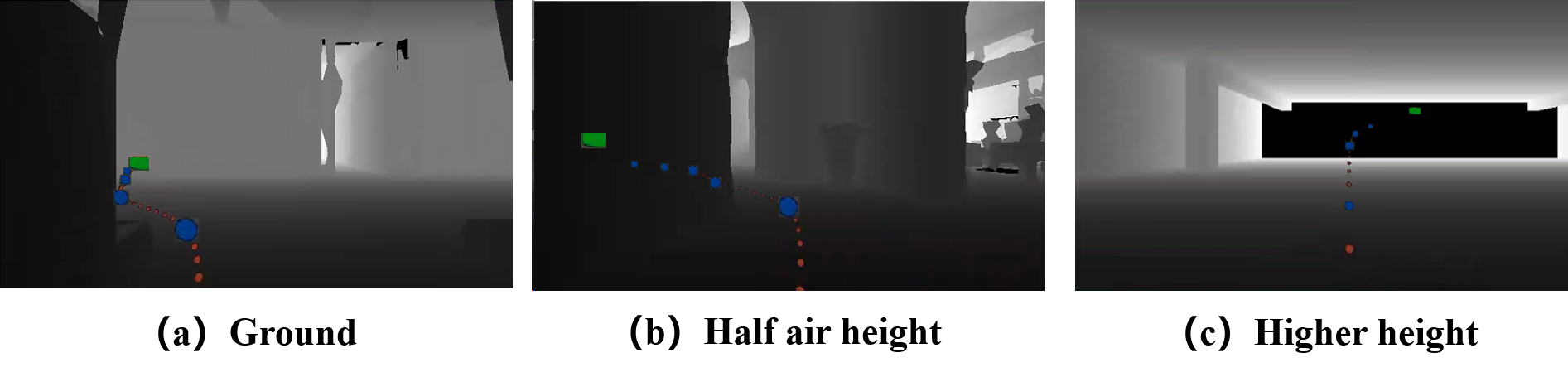}
      \caption{Navigation trajectories in the simulation environment when the goal is at different heights.}
      \label{height}
\vspace{-0.2cm}
\end{figure}

It should be noted that to validate our approach can extend the iPlanner's land navigation to 3D air navigation, we randomly adjusted the goal point positions in the dataset, allowing goals to appear on either land or in the air. The robot had to follow the objective function in Equation \ref{total} to approach the goal with minimal energy consumption. Some of our experimental results are shown in Fig. \ref{height}, where regardless of the goal's height, our method consistently generated collision-free trajectories close to the goal based on the key points path.

\subsection{Real World Indoor Multi-Scene Experiments}
This experiment will evaluate the performance of our lightweight navigation method. We have set up two navigation tasks with different goal positions: one is set on the land, and the other is set in mid air. For each task, we will compare the path length, time, and energy with existing methods, and the results are shown in Table \ref{table3}.

\begin{table}[ht]
\caption{Performance comparison of LABV with existing methods in different task.}
\label{table3}
\centering
\begin{tabular}{ccccc}
\hline
\multirow{2}{*}{Task} & \multirow{2}{*}{Method}  & \multirow{2}{*}{Length(Land/Air)} & \multirow{2}{*}{Time(s)} & \multirow{2}{*}{Energy(J)}\\
 &  & & & \\
\hline
\multirow{3}{*}{\shortstack{Land\\Navigation}} & Yang's \cite{iPlanner} & \textbf{7.5m}/0m & 13.7 &1161.44 \\
 & Zhou's \cite{Ego}  &10.7m/0m & 19.2 & 1544.41 \\
 & Ours &8.6m/0m &\textbf{11.8} &\textbf{1008.70} \\
\hline
\multirow{3}{*}{\shortstack{Land-air\\Navigation}} & Yang's \cite{iPlanner} & - / - & - & - \\
& Zhou's \cite{Ego} &3.2m/5.8m  & 22.5 & 17225.8 \\
 & Ours &7.2m/\textbf{1.4m} & 28.7 & \textbf{11010.7}\\
\hline
\end{tabular}
\vspace{-0.2cm}
\end{table}

\begin{figure}[thpb]
      \centering
    \includegraphics[width=1\linewidth]{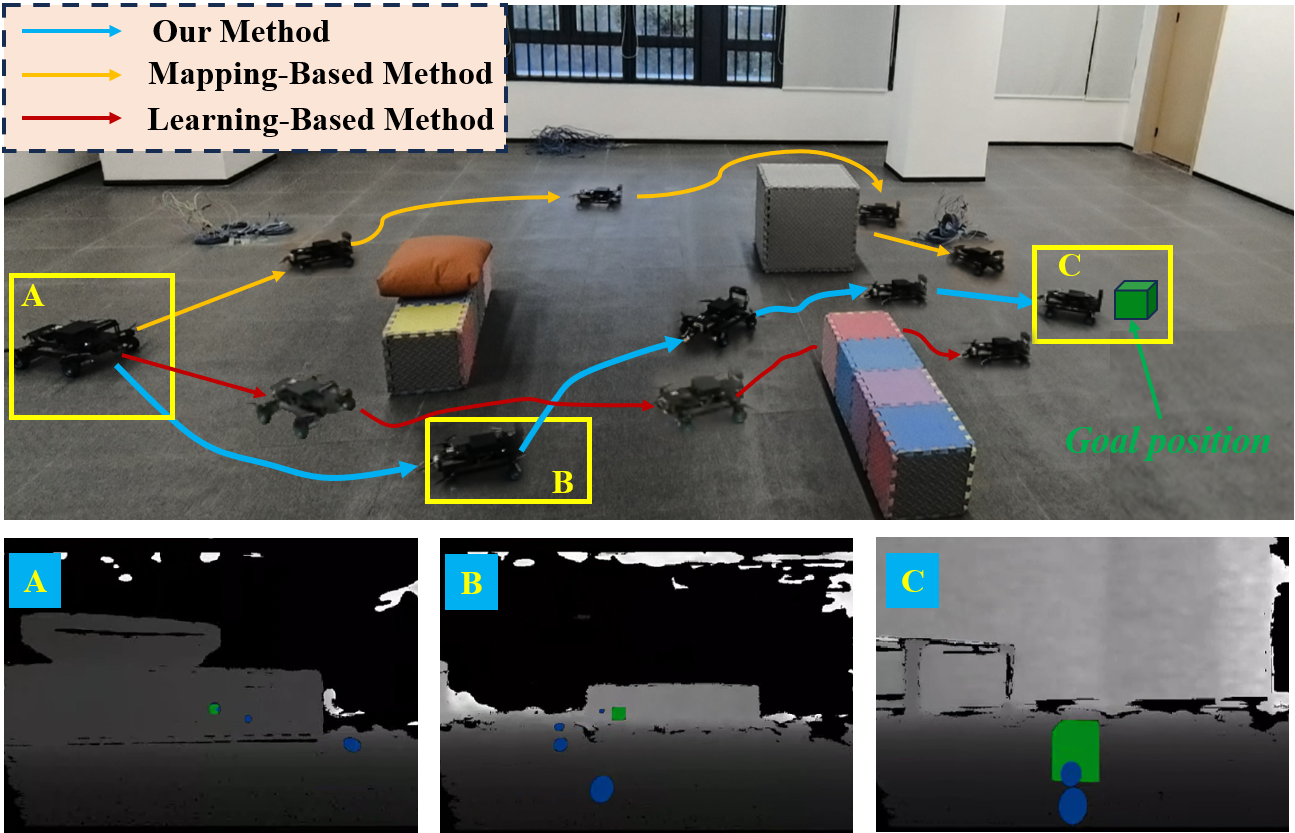}
      \caption{Real-world experiment on different methods to land navigation. Three local planning events (A-C) of our proposed method along the path with corresponding depth images are shown with the estimated key points.}
      \label{compare2d}
\end{figure}

The first task is shown in Fig. \ref{compare2d}. The mapping-based approach proposed by Zhou et al.\cite{Ego} generates more conservative trajectories when the goal lies on the land. This suboptimal path planning results from inherent safety constraints imposed by limited sensor perception in unknown environments. In contrast, although the learning-based method developed by Yang et al.\cite{iPlanner} predicts a shorter theoretical path, experimental observations reveal collisions due to insufficient environmental adaptation in currently untrained scenarios. Furthermore, hardware limitations in our onboard computer (i.e., absence of GPU acceleration) introduce processing latency in image streams. This necessitates reducing trajectory point transmission frequency to synchronize with the controller, consequently decreasing navigation velocity and prolonging overall mission duration.

As indicated in Table \ref{table3}, our two-stage methodology demonstrates superior performance: the first stage ensures real-time operation by generating global key points through GKPN, while the second stage enables intelligent local decision-making. Our architecture achieves balanced path length optimization while attaining the shortest navigation duration (11.8s) and minimal energy consumption (1008.7J) among comparative methods.

\begin{figure}[thpb]
      \centering
    \includegraphics[width=1\linewidth]{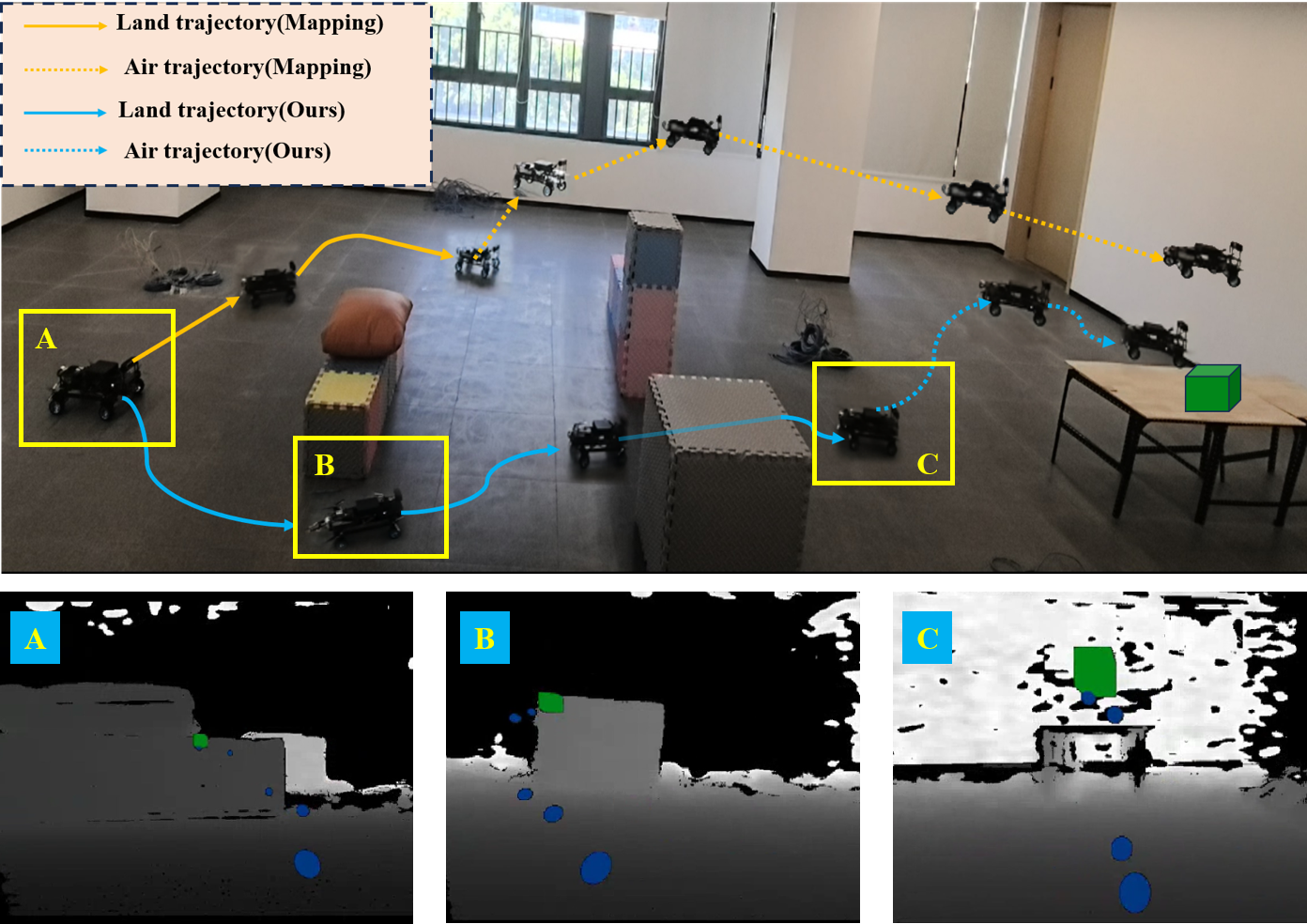}
      \caption{Experiments on land-air switching scenarios with different methods. Three local planning events (A-C) of our method along the path with corresponding depth images are shown with the key points.}
      \label{compare3d}
\end{figure}

The second task is shown in Fig. \ref{compare3d}. The iPlanner approach has limitations in air navigation due to land constraints. When encountering an air goal beyond the sensor coverage, mapping-based methods typically activate air mode in advance and extend the air trajectory to safely approach the goal. This prolonged flight leads to higher energy consumption, as shown in Table \ref{table3}. In contrast, our method initially generates key points on the land and only transitions to air mode when the robot is close to the goal and the key points reach a certain height. By minimizing unnecessary air displacement, our method reduces the flight distance by 4.4m, achieving a 35\% reduction in overall energy consumption while maintaining navigational efficiency.

\section{CONCLUSIONS}

This work presents a two-stage lightweight framework for land-air bimodal navigation that synergizes learning-based global prediction with mapping-based local planning to address the critical trade-off between computational efficiency and path optimality. The proposed GKPN enhances obstacle perception through edge feature extraction and lightweight attention mechanisms. Experimental results validate significant improvements in energy efficiency (35\% reduction) and parameter count (14\% decrease) compared to conventional approaches. Our implementation on a GPU-free LABR platform further demonstrates practical viability under computational constraints, achieving seamless mode transitions between land and air navigation. 




\section*{ACKNOWLEDGMENT}

 This work was supported by GuangMing Laboratory, under project No.23501002.

\bibliographystyle{ieeetr}
\bibliography{root}

\end{document}